%% file: main.tex
\documentclass[10pt,twocolumn,letterpaper]{article}

\usepackage[pagenumbers]{cvpr} 
\usepackage{multirow}
\usepackage{bbding}

\input{preamble}
\definecolor{cvprblue}{rgb}{0.21,0.49,0.74}
\usepackage[pagebackref,breaklinks,colorlinks,citecolor=cvprblue]{hyperref}

% ...

\definecolor{hookersgreen}{rgb}{0.0, 0.44, 0.0}

\def\paperID{236} 
\def\confName{3DV\xspace}
\def\confYear{2025\xspace}

\title{A2-GNN: Angle-Annular GNN for Visual Descriptor-free Camera Relocalization}

\author{Yejun Zhang$^{1}$ \ \ \ Shuzhe Wang$^{1}$ \ \ \ Juho Kannala$^{1,2}$ \\ 
$^{1}$Aalto University  \ \ \ $^{2}$University of Oulu \\
{\tt\small firstname.lastname@aalto.fi}
}

\begin{document}
\maketitle
\input{sec/0_abstract}    
\input{sec/1_intro}
\input{sec/2_related_work}
\input{sec/3_method}

\input{sec/4_experiments}
\input{sec/5_conclusion}
{
    \small
    \bibliographystyle{ieeenat_fullname}
    \bibliography{main}
}
\input{sec/X_suppl}

\end{document}

%% file: preamble.tex
\usepackage[dvipsnames]{xcolor}

\definecolor{azure}{rgb}{0.0, 0.5, 1.0}
\definecolor{frenchblue}{rgb}{0.0, 0.45, 0.73}
\definecolor{forestgreen(traditional)}{rgb}{0.0, 0.27, 0.13}
\definecolor{mygreen}{rgb}{0.09, 0.45, 0.27}
\definecolor{myblue}{rgb}{0.2383,0.5195,0.7734}
\definecolor{mygreen}{rgb}{0.6445,0.9297,0.0039}
\definecolor{darklavender}{rgb}{0.45, 0.31, 0.59}
\definecolor{americanrose}{rgb}{1.0, 0.01, 0.24}
\definecolor{pigblue}{rgb}{0.2, 0.2, 0.6}
\definecolor{blue(ryb)}{rgb}{0.01, 0.28, 1.0}
\definecolor{amethyst}{rgb}{0.6, 0.4, 0.8}
\definecolor{deepmagenta}{rgb}{0.8, 0.0, 0.8}
\definecolor{carminered}{rgb}{1.0, 0.0, 0.22}
\definecolor{iris}{rgb}{0.35, 0.31, 0.81}

\newcommand{\reals}{{\mbox{\bf R}}}

\def \reals    {{\mathbb R}}

\def\ccalM{{\ensuremath{\mathcal M}}}

\def\ccalP{{\ensuremath{\mathcal P}}}
\def\ccalQ{{\ensuremath{\mathcal Q}}}

\def\ccal0{{\ensuremath{\mathcal 0}}}
\def\bbK{{\ensuremath{\mathbf K}}}

\def\bbW{{\ensuremath{\mathbf W}}}

\def\bbb{{\ensuremath{\mathbf b}}}
\def\bbc{{\ensuremath{\mathbf c}}}

\def\bbe{{\ensuremath{\mathbf e}}}
\def\bbf{{\ensuremath{\mathbf f}}}
\def\bbg{{\ensuremath{\mathbf g}}}

\def\bbk{{\ensuremath{\mathbf k}}}

\def\bbm{{\ensuremath{\mathbf m}}}

\def\bbp{{\ensuremath{\mathbf p}}}
\def\bbq{{\ensuremath{\mathbf q}}}

\def\bbw{{\ensuremath{\mathbf w}}}

\def\bbv{{\ensuremath{\mathbf v}}}

\def\bb0{{\ensuremath{\mathbf 0}}}

%% file: sec/0_abstract.tex
\begin{abstract}
Visual localization involves estimating the 6-degree-of-freedom (6-DoF) camera pose within a known scene. 
A critical step in this process is identifying pixel-to-point correspondences between 2D query images and 3D models. 
Most advanced approaches currently rely on extensive visual descriptors to establish these correspondences, facing challenges in storage, privacy issues and model maintenance.
Direct 2D-3D keypoint matching without visual descriptors is becoming popular as it can overcome those challenges.
However, existing descriptor-free methods suffer from low accuracy or heavy computation.
Addressing this gap, this paper introduces the Angle-Annular Graph Neural Network (A2-GNN), 
a simple approach that efficiently learns robust geometric structural representations with annular feature extraction.
Specifically, this approach clusters neighbors and embeds each group's distance information and angle as supplementary information to capture local structures.
Evaluation on matching and visual localization datasets demonstrates that our approach achieves state-of-the-art accuracy with low computational overhead among visual description-free methods. Our code will be released on \url{https://github.com/YejunZhang/a2-gnn}.
\end{abstract}

%% file: sec/1_intro.tex
\section{Introduction}
\label{sec:intro}

\begin{figure}[!t]
    \centering
    \includegraphics
    [width=0.95\linewidth] {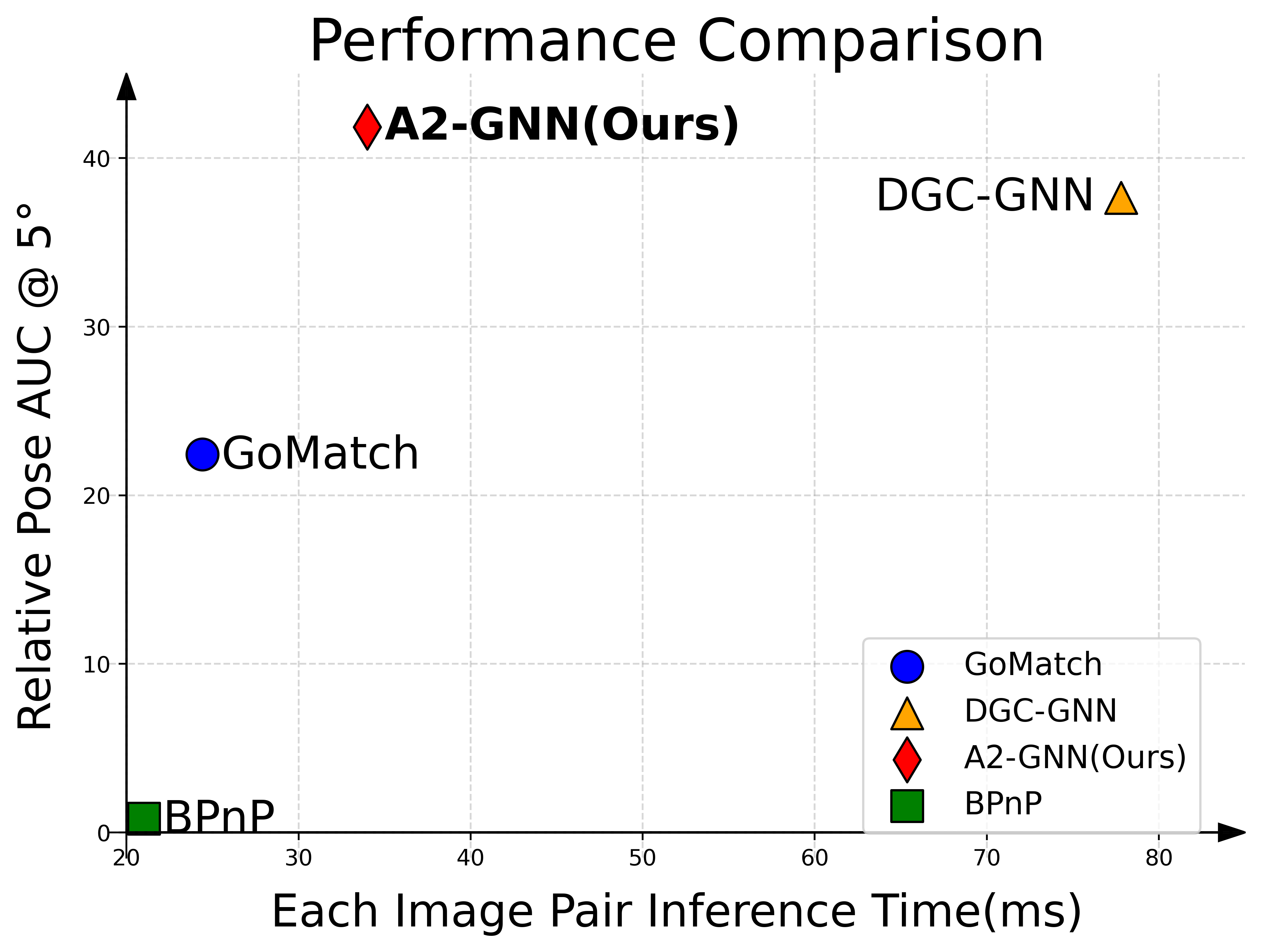}
    \caption{Matching Accuracy and Efficiency Comparisons for descriptor-free methods. Compared with GoMatch~\cite{zhou2022geometry} and DGC-GNN~\cite{wang2024dgc}, our A2-GNN learns effective and accurate 2D-3D matching.
    }
    \label{fig: time}    
\end{figure}

Visual localization aims to estimate the camera pose of a query image within a pre-built 3D environment. 
It is essential in computer vision applications such as Structure-from-Motion (SfM)~\cite{schonberger2016structure,snavely2006photo,wu2013towards},
Augmented Reality (AR)~\cite{carmigniani2011augmented,billinghurst2015survey}, and Simultaneous Localization and Mapping (SLAM)~\cite{cadena2016past,fuentes2015visual,mur2017orb} . 
Classical visual localization approaches~\cite{sattler2011fast,sattler2012improving, sattler2017large,sarlin2019coarse,wang2021continual,wang2024hscnet++}  
use visual descriptors to build the 2D-3D correspondences.
Once the correspondences are established, a PnP-RANSAC solver~\cite{zuliani2009ransac} is applied to estimate the 6-DoF camera pose.
 
Among all camera pose estimation approaches, visual descriptor-based methods~\cite{sarlin2019coarse,sarlin2020superglue} achieve state-of-the-art (SOTA) performance. 
However, utilizing visual descriptors presents several challenges, including substantial storage requirements, privacy risks, and maintenance complexity~\cite{zhou2022geometry,wang2024dgc,chelani2021privacy,pan2023privacy}. 
Storing high-dimensional descriptors for each keypoint typically demands significant storage capacity.
Additionally, studies~\cite{pan2023privacy,chelani2021privacy} have demonstrated that visual descriptors can be exploited to recover images, raising privacy concerns.
Moreover, maintaining the 3D model becomes complicated when integrating new descriptors or points into existing point clouds~\cite{dusmanu2021cross}.

To address the challenges associated with visual descriptor-based methods, several researchers have developed visual descriptor-free approaches~\cite{campbell2020solving,zhou2022geometry,wang2024dgc}. 
These methods establish 2D-3D correspondences without relying on visual descriptors, instead leveraging geometric information from keypoints for matching. 
While these approaches demonstrate reasonable performance, their primary limitation lies in the inherently limited geometric information, which lacks the richness and detail provided by visual descriptors.  GoMatch~\cite{zhou2022geometry} attempts to learn the geometric-only representation by utilizing graph neural networks and incorporating outlier rejection, achieving reasonable performance. 
However, there is still a noticeable performance gap compared to descriptor-based methods. 
DGC-GNN~\cite{wang2024dgc} narrows this gap by integrating color and more geometric clues, along with global-to-local clustering. 
Despite these improvements, DGC-GNN suffers from heavy computational demands, particularly due to the multiple clustering operations required.

To achieve accurate camera pose estimation with light computation,  we reconsider the way of encoding geometric information. In previous descriptor-free approaches~\cite{zhou2022geometry,wang2024dgc}, transformers~\cite{vaswani2017attention} have been employed, utilizing self-attention and cross-attention mechanisms to enhance feature representation. \
The self-attention mechanism captures local geometric structures within neighborhoods, while cross-attention facilitates the exchange of geometry information between 2D keypoints and 3D point clouds. 
The extraction of local geometry through self-attention is particularly crucial, as learning high-quality feature representations is essential for effective information exchange in the following cross-attention layer.
Previous methods~\cite{zhou2022geometry,wang2024dgc} utilize max-pooling to extract local geometry from neighbors. 
However, this operation neglects neighbor structural information as most of neighbor information is discarded.
Inspired by CLNet~\cite{zhao2021progressive}, we propose \textbf{A}ngle-\textbf{A}nnular \textbf{G}raph \textbf{N}eural \textbf{N}etwork, or \textbf{A2-GNN}, 
which efficiently extracts local structural geometric information for 2D-3D matching without relying on visual descriptors. Our network processes sparse points from query images and point clouds as inputs, delivering accurate correspondences with minimal computational overhead.
We first construct a local graph for each point by connecting it to its neighboring points based on their distances. These neighboring points are then clustered into close, middle, and remote groups. Local structural information is extracted by processing these grouped neighbors separately. Besides, angular information between the point and its neighbors is encoded similarly to further enhance the local geometric representation. By encoding these geometric cues, our method effectively distinguishes between structurally similar yet distinct points in both 2D and 3D spaces.
In addition, we also adjust outlier rejection input to position information rather than the feature representation, as explicit position information can provide consensus in epipolar geometry to remove outliers.
In summary, our paper makes the following contributions:
\begin{itemize}
  \item We introduce a novel local graph neighbor aggregation method for direct 2D-3D matching without visual descriptors. This method embeds geometric information in a hybrid manner and improves the accuracy of spare 2D-3D matching.
  \item We adjust outlier rejection input to position information, as pure position information can provide robust geometry constraint. 
  \item Our method outperforms previous descriptor-free methods in matching and visual localization tasks with considerable efficiency.
\end{itemize}

%% file: sec/2_related_work.tex
\begin{figure*}[!t]
    \centering
    \includegraphics
    [width=0.8\linewidth] {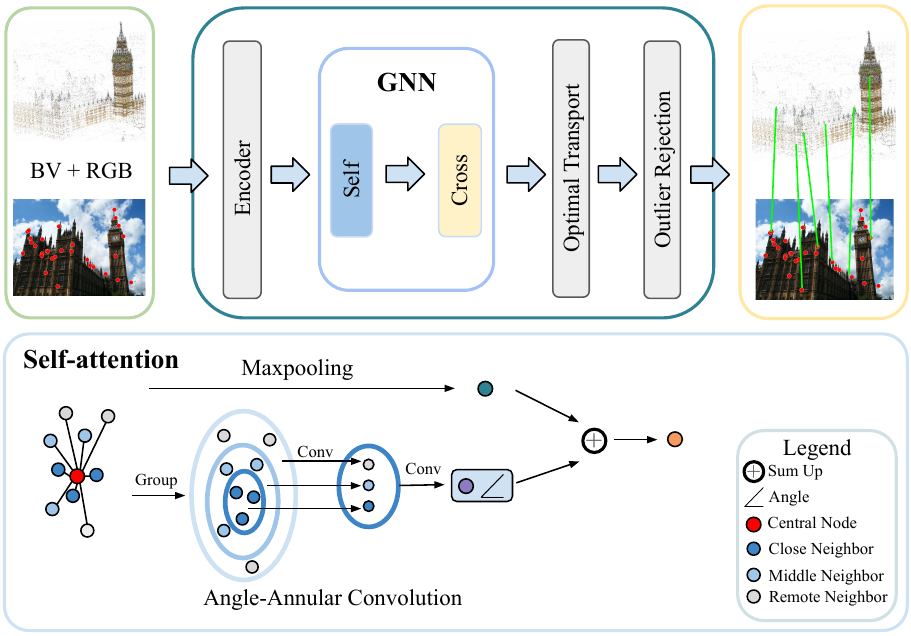}
    \caption{Architecture Overview. The bearing vector (BV) and RGB information from the query image and 3D point cloud are first processed through an encoder to generate high-dimensional features. These features are then used to construct the graph nodes. In the self-attention layer, the angle-annular convolution is employed to extract discriminative geometric information from the neighboring points. After the GNNs, these enhanced features are used to establish initial correspondences via optimal transport. Outlier rejection is then applied to eliminate erroneous correspondences, resulting in a final set of accurate correspondences.}
    \label{fig:network}    
\end{figure*}

\section{Related work}
\label{sec:related work}

\noindent
\textbf{Structure-Based Localization.}
Structure-based visual localization has achieved high accuracy by leveraging a pre-built 3D map of the environment, which is crucial for establishing correspondences between 2D image pixels and 3D point clouds. Classical methods~\cite{sattler2011fast,sattler2012improving} relied on Structure-from-Motion (SfM) models to match features between query images and 3D points. In contrast, recent approaches have incorporated learning-based techniques, significantly enhancing image retrieval~\cite{arandjelovic2016netvlad,ge2020self,gordo2017end,revaud2019learning}, feature extraction~\cite{detone2018superpoint,revaud2019r2d2, dusmanu2019d2,tyszkiewicz2020disk}, and keypoint matching~\cite{arandjelovic2012three,sarlin2020superglue,lindenberger2023lightglue} in the structure-based localization pipeline. 

Image retrieval is a vital component of structure-based localization, as it identifies similar or overlapping images from a large database to narrow down the search space of the 3D points.
NetVLAD~\cite{arandjelovic2016netvlad} integrates a VLAD-like layer in CNNs and optimizes local descriptor extraction and aggregation, enhancing image retrieval accuracy.
Recent works~\cite{keetha2023anyloc,Izquierdo_CVPR_2024_SALAD} leverage powerful pretrained models to further improve retrieval accuracy.
Feature extraction methods~\cite{detone2018superpoint,revaud2019r2d2,dusmanu2019d2} 
extract robust visual representations from raw images, enabling more reliable keypoint detection and description. The integration of neural networks has further enhanced keypoint matching accuracy, particularly with advanced methods like SuperGlue~\cite{sarlin2020superglue}, which employs graph neural networks to refine the matching process. Recent advancements~\cite{shi2022clustergnn,lindenberger2023lightglue,jiang2024omniglue} continue to push the boundaries of accuracy and efficiency in this domain, making structure-based visual localization increasingly robust and scalable for real-world applications. 
In addition, detector-free methods~\cite{sun2021loftr,chen2022aspanformer,wang2024efficient} input images directly for matching and achieve significant improvements in accuracy but encounter challenges in heavy computation.

\noindent
\textbf{Descriptor-free Visual Localization.}
BPnPNet~\cite{campbell2020solving} makes significant progress in 2D pixel to 3D point cloud matching without relying on visual descriptors, introducing an end-to-end trainable matcher for cross-modal point matching. It also introduces the use of bearing vectors as a 2D keypoint representation, effectively bridging the 2D-3D gap and performing well in outlier-free scenarios. GoMatch~\cite{zhou2022geometry} further develops this concept, utilizing bearing vectors to represent both 2D and 3D keypoints for geometric-only matching. It employs SuperGlue-style~\cite{sarlin2020superglue} self- and cross-attention mechanisms to establish initial correspondences and integrates PointCN~\cite{yi2018learning} to remove outliers from the initial matches. DGC-GNN~\cite{wang2024dgc} further improves matching accuracy by leveraging additional RGB information, multiple geometric cues, and a global-to-local feature embedding strategy. It clusters points to create a global geometric graph, enhancing feature matching across clusters by encoding Euclidean and angular relations. However, this approach comes with a trade-off: its inference time is three times longer than that of GoMatch due to the iterative point clustering process.

\noindent
\textbf{Scene Compression.}
Scene compression for visual localization involves two primary approaches: map compression and descriptor compression. Map compression~\cite{camposeco2019hybrid,cao2014minimal,dymczyk2015gist,yang2022scenesqueezer,li2010location,laskar2024differentiable} reduces map size by pruning indistinguishable points,
while descriptor compression~\cite{jegou2010product,sattler2015hyperpoints,ke2004pca,dong2023learning,yang2022scenesqueezer,laskar2024differentiable} minimizes descriptor size while maintaining their essential properties for accurate matching.  
Scene coordinate regression~\cite{brachmann2021visual,brachmann2023accelerated, li2020hierarchical, wang2024hscnet++,wang2021continual, wang2024glace} uses neural networks to compress scene models and learns a compact representation.
It predicts scene coordinates directly from 2D images without visual descriptors, but faces challenges in generalizing to new scenes.
Hybrid methods~\cite{camposeco2019hybrid,yang2022scenesqueezer,laskar2024differentiable} combine both map and descriptor compression to achieve more compact representations. However, while scene compression techniques can save storage space, they do not fully address privacy concerns or the complexities associated with descriptor maintenance. Our method, which avoids the use of visual descriptors altogether, can also be integrated with scene compression techniques to further reduce the model size.

\noindent
\textbf{Privacy-preserving visual localization.}
As demonstrated in~\cite{pittaluga2019revealing}, image details can be recovered from descriptors. 
This raises significant data privacy concerns, as descriptors can be leaked during transmission across devices.
To address this concern, privacy-preserving visual localization methods~\cite{speciale2019privacy1,speciale2019privacy2,shibuya2020privacy,pan2023privacy,moon2024efficient} have been developed.
These methods represented 2D/3D points as lines or other geometric representations to obscure spatial details. However, they suffer from accuracy degradation, high storage requirements, and maintenance complexity.

Other privacy-preserving methods~\cite{do2022learning,pietrantoni2023segloc} take alternative strategies. SLD~\cite{do2022learning} identifies distinctive points as scene landmarks for camera localization.
SegLoc~\cite{pietrantoni2023segloc} utilizes semantic segmentation to create feature representations optimized for privacy and localization performance. Nevertheless, these methods rely on extensive priors or labels and limit generalizability in other scenarios. 


%% file: sec/3_method.tex
\section{Method}
\label{sec:method}

\subsection{Problem setting}

Let $p \in \reals^{2} $ denote a 2D point and $q \in \reals^{3} $ denote a 3D point.
Assume that a 2D image $P$ contains $M$ keypoints, and a 3D point cloud $Q$ contains $N$ keypoints.
We denote the sets of keypoints in 2D and 3D cases as $ \ccalP = \{p_i \mid i=1,\ldots,M\} $ and $ \ccalQ = \{ q_j \mid j = 1, \ldots, N\} $, respectively.
Our task is to predict a set of correspondences $\ccalM_{p,q} 
$ between 2D keypoints and 3D point clouds.

\noindent
\textbf{Keypoint Representation.}
Following GoMatch \cite{zhou2022geometry}, we utilize bearing vectors as keypoint representation, as it can bring 2D keypoints and 3D point clouds into the same modality.
For 2D keypoints, bearing vectors $\bbb_\bbp$ remove the effect of camera intrinsic $\bbK$
by the following equation:
\begin{equation}
    [\bbb_\bbp^\top, 1]^\top = \mathbf{K}^{-1}[u, v, 1]^\top ,
\end{equation} 
where $(u, v)$ represents 2D keypoint pixel coordinates and $\bbb_\bbp \in \reals^2$.  
The bearing vector brings pixel coordinates into the corresponding camera ray by connecting the camera center and pixel coordinates.
For 3D points, 
the bearing vector $\bbb_\bbq$ transforms its world coordinates $\bbp^\bbw$ to the camera ray by the following equation:
\begin{align}
    \bbp^\bbc & =  \mathbf{R}  \bbp^\bbw + \mathbf{t} \\
    [ \bbb_\bbq^\top, 1]^\top & =  \bbp^\bbc / {p^c_z},
\end{align}
where $\bbp^\bbc \in \reals^3 $ is its the camera coordinate, $p'_z$ is its $z$ coordinate and $\bbb^\bbq \in \reals^2$. 
The bearing vector first transforms 3D points from the world coordinates to the camera coordinates. 
Similar to the image plane, we take plane $z=1$ and connect the camera center to its camera coordinates.
The camera ray is obtained by connecting the camera center and the point where the line intersects the plane.

\subsection{Network Architecture}
We give an overview of the proposed Angle-Annular convolution Graph Neural Network, shortened as A2-GNN in Fig.~\ref{fig:network}. 
The architecture includes the following modules: feature encoder,  angle-annular geometric feature extraction, optimal transport, and outlier rejection.
The encoder transforms positional information and RGB color from low-dimensional inputs into high-dimensional features.
These features are then processed by A2-GNN to extract local geometric relations from neighboring points. Initial correspondences are established using optimal transport, and low-confidence matches are subsequently removed by the outlier rejection module.

\subsubsection{Feature Encoder}
\textbf{Feature Encoder.} We use a ResNet-style encoder~\cite{he2016deep,campbell2020solving,zhou2022geometry} to extract both position and color features directly from 2D kepoints or 3D point clouds, denoted as $\mathcal{F}_{b}$ and $\mathcal{F}_{c}$. The feature encoder encodes the bearing vector and RGB color separately, from low dimension vectors to high-dimensional (e.g., $d$=128) vectors. The point features $\bbf_p \in \reals^{N\times d}$ from query images and $\bbf_q \in \reals^{M\times d}$ from point clouds can be computed as:
\begin{subequations}
    \begin{align}
    \bbf_\bbp & = \mathcal{F}_{b}(\bbb_\bbp) + \mathcal{F}_{c} ( \bbc_\bbp) \\
    \bbf_\bbq & = \mathcal{F}_{b}(\bbb_\bbq) + \mathcal{F}_{c} (\bbc_\bbq),
\end{align}
\end{subequations}
where $\bbc_\bbp \in \reals ^ 3, \bbc_\bbq \in \reals ^ 3$ are RGB color of points from the query images and point clouds, respectively.

\subsubsection{A2-GNN} 

Graph Neural Networks enhance feature representation by aggregating information from neighboring nodes, which is utilized for both images and point clouds.
This section introduces the Angle-Annular Convolution Aggregation method, self-attention and cross-attention mechanisms.

\noindent
\textbf{Geometric Local Feature.}
After feature extraction, we construct local graphs for each keypoint based on its neighboring points. We consider the local graphs as a self-attention mechanism to extract local geometric features, thereby enhancing the representation and capturing the contextual relationships among the keypoints. Following previous works~\cite{zhou2022geometry,wang2024dgc}, edges $\mathcal{E}$ are constructed in Euclidean space by connecting each node to its $k$ nearest neighbors. The edge feature $\bbe_{ij} \in \reals ^{k \times 2d}$ between a node $\bbf_i$ and its neighbors $\bbf_{ij}$ is defined as:
\begin{equation}
    \bbe_{ij} = \text{cat}[ \bbf_i, \bbf_i - \bbf_{ij}], j = 1,2, \ldots k,
\end{equation}
where $\text{cat}[\cdot,\cdot]$ denotes concatenation. 
Then maxpooling operation is used to extract local information from neighbour nodes.
The feature $\bbf_{max} \in \reals^{1 \times d}$ can be updated using following equation:
\begin{equation}
    ^{(t+1)}\bbf_{max} = \max_{(i,j)\in \mathcal{E}} h_{\theta}( ^{(t)}\bbe_{ij}),
\end{equation}
where $h_{\theta}$ represents a linear layer followed by instance normalization~\cite{ulyanov2016instance} and LeakyReLU, and $\text{max}(\cdot)$ is element-/channel-wise maxpooling.

However, the maxpooling operation disregards the inherent structural relationships between nodes in the graph, as each graph node only retains the maximum value from its neighbors. We observe that if a pair of points are correct correspondences between the keypoints from the image and the point cloud, they should exhibit similar structures or geometric patterns within their respective neighborhoods. 
In order to dig more inherent structural geometry information between the node and its neighbors, we introduce \textbf{Angle-Annular convolution}.
First, we use annular convolution inspired by CLNet~\cite{zhao2021progressive} to capture the relationships of neighbors in a grouped manner. 
As shown in Fig.~\ref{fig:network} bottom, $k$ neighboring nodes are determined by ranking the Euclidean distances. Those $k$ nodes are then divided into $g$ groups where each group contains $\frac{k}{g}$ nodes.
We use convolution layers followed by one Batch Normalization layer~\cite{ioffe2015batch} with ReLU to process the grouped graphs. 
This convolution utilizes distance information indirectly by dividing nodes based on the distance between node and its neighbors.
The annular feature $\bbf_{ann} \in \reals^{1 \times d}$ can be formulated as:
\begin{equation}
    ^{(t+1)} \bbf_{ann}= g_2(g_1(^{(t)}\bbe_{ij})),
\end{equation}
where $g_1(\cdot)$ and $g_2(\cdot)$ denotes the convolution layers with $1 \times \frac{k}{g}$ kernels and $1\times g$ kernels.
 
\begin{figure}[h]
    \centering
    \includegraphics
    [width=0.8\linewidth] {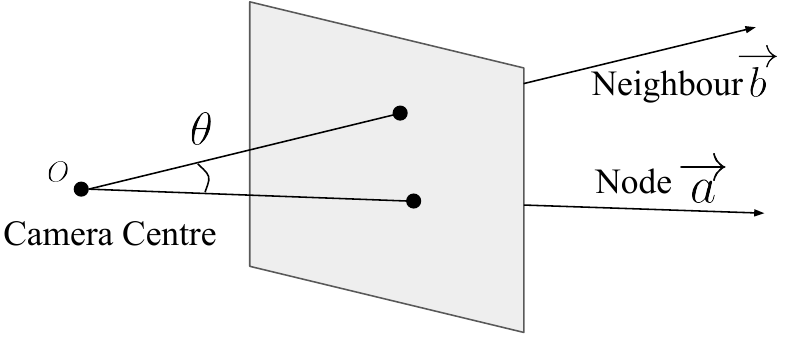}
    \caption{Illustration of angle embedding. The angle embedding between node and its neighbor is added to enhance feature representation.}
    \label{fig:angle}    
\end{figure}

Annular convolution efficiently captures structural information, but its effectiveness is limited by the variability in spatial distances between nodes and their neighbors, with some neighbors being adjacent and others more distant. Inspired by~\cite{qin2022geometric,wang2024dgc}, we incorporate geometric encoding that includes both distance and angle embeddings. The distance embedding is implicitly applied in constructing node neighborhoods. As illustrated in Fig \ref{fig:angle}, angle embedding between a node and its neighbors can also be incorporated. The cosine of the angle $\theta$, where $\theta \in (0, \pi)$, is used as the angle embedding.
The cosine value can be computed as:
\begin{equation}
    \text{cos} (\theta )= \frac{\overrightarrow{a} \cdot \overrightarrow{b}}{\| \overrightarrow{a} \| \| \overrightarrow{b} \|},
\end{equation}
where $\text{cos} (\cdot)$ is cosine function. After getting each angle embedding within its neighbors, we get angle edge $\mathcal{E}_{ang} \in \reals ^{k \times 1}$.
Here, we use annular convolution again to get angle feature $\bbf_{ang} \in \reals ^ {1 \times d}$:
\begin{equation}
    ^{(t+1)} \bbf_{ang} = g_4(g_3(\text{cos}(^{(t)} \mathcal{E}_{ang}))),
\end{equation}
where $g_3(\cdot)$ and $g_4(\cdot)$ are similar convolutions as $g_1(\cdot)$ and $g_2(\cdot)$ but with independent parameters.
After getting the angle feature, the annular-angle features are computed by merging the angle feature and angular feature as follows: $\bbf_{aa} = \bbf_{ann} + \bbf_{ang}$. The final feature $\bbf_{self}$ can be updated twice
and the final self-attention feature $\bbf_{self}$ are obtained as: 
\begin{align}
    \bbf_{self} 
    & = h_{\theta 1}(\text{cat}[^{(0)}\bbf,^{(1)}\bbf_{max},^{(2)}\bbf_{max}]) \notag \\
    & + h_{\theta 2}(\text{cat}[^{(0)} \bbf,^{(1)}\bbf_{aa},^{(2)}\bbf_{aa}],
\end{align}
where $h_{\theta 1}$ and $h_{\theta 2}$ denote a linear layer followed by instance normalization~\cite{ulyanov2016instance} and LeakyReLU.

\noindent
\textbf{Cross-attention layer.} 
Following~\cite{zhou2022geometry}, we apply cross-attention to enhance feature representation. Each graph node from the 2D is connected with every node from the point cloud. Specifically, given a node feature $\bbf_i$ from one modality and features $\bbg_j$ from the other modality, we form the query, keys and values according to 
$\boldsymbol{\mathsf{q}}_i = \bbW_\mathsf{q} \bbf_i, \bbk_j = \bbW_k \bbg_j$ and $\bbv_j = \bbW_v \bbg_j$, where $\bbW_s,\bbW_k,\bbW_v \in \reals^{d \times d}$ are learned parameters. We update feature $\bbf_i$ by following formulation:
\begin{align}
    \bbm_i & = \sum_j \alpha_{ij} \bbv_j \\
    ^{(t+1)} \bbf_{i} & = ^{(t)} \bbf_i + \text{MLP}(\text{cat}[\boldsymbol{\mathsf{q}}_i,\bbm_i]),
\end{align}
where attention weights $\alpha = \text{softmax}(\boldsymbol{\mathsf{q}}_j^\top \bbk_j / \sqrt{d})$.

\subsubsection{Optimal Transport}
After obtaining the 2D and 3D enhanced features $f^G_p$ and $f^G_q$ from the geometric embedding, optimal transport is employed to assign initial correspondences, denoted as $\ccalM_{init}$. We first compute the cost matrix $M \in \reals^{M \times N}$ using the $\text{L}2$ distance 
between those two feature sets. To handle unmatched points, we extend $M$ to $M' \in \reals^{(M+1) \times (N+1)}$ by adding an additional row and column as dustbins. The differentiable Sinkhorn algorithm~\cite{cuturi2013sinkhorn,sinkhorn1967concerning} is then applied to solve the optimal transport problem. Finally, the initial correspondences $\ccalM_{init}$ are obtained by removing the dustbins from $M'$ and performing a mutual nearest neighbor check.

\subsubsection{Outlier Rejection}
After the optimal transport layer, the initial correspondences $\ccalM_{init}$ still contain outliers. GoMatch~\cite{zhou2022geometry} and DGC-GNN~\cite{wang2024dgc} utilize an outlier classifier~\cite{yi2018learning} to predict scores indicating the probability of a correspondence being an inlier. The final correspondences $\ccalM_{final}$ are obtained if the probability is over the threshold $t$ (0.5 in experiments).
The method directly inputs enhanced geometric features into the classifier. However, it brings limitations as the enhanced features are primarily designed for the optimal transport layer and are mixed together with the RGB embedding, making it difficult to extract the epipolar geometry constraints necessary for accurate outlier rejection. Earlier learning-based outlier rejection methods~\cite{yi2018learning,zhang2019learning,zhao2021progressive} leveraged keypoint locations as input, which naturally encode epipolar geometry constraints. To address this issue, we propose using bearing vectors as input for the outlier rejection process, allowing the model to utilize geometric information better and improve the accuracy of inlier detection. 

\subsection{Training Loss}
Following GoMatch~\cite{zhou2022geometry} and DGC-GNN~\cite{wang2024dgc}, the loss function $\mathcal{L}$ consists of a matching loss $\mathcal{L}_{m}$ and an outlier rejection loss $\mathcal{L}_{or}$.
The matching loss $\mathcal{L}_{m}$ is designed to minimize the negative log-likelihood of the matching scores. 
The matching loss is presented as:
\begin{align}
    \mathcal{L}_{m} &= - \frac{1}{N_m} (\sum_{(i, j) \in \mathcal{M}_{gt}} \log \tilde{\text{P}}_{ij} +  \sum_{i \in \mathcal{U}_{q}} \log \tilde{\text{P}}_{i(N + 1)} \notag \\
    &+ \sum_{j \in \mathcal{U}_{d}} \log \tilde{\text{P}}_{(M + 1)j} ), 
\end{align}
where $\tilde{\text{P}}$ means matching score, 
$\mathcal{M}_{gt}$ denotes ground truth matches, $\mathcal{U}_{q}$ represents unmatched query keypoints and $\mathcal{U}_{d}$ is unmatched database 3D points. $N_m$ refers to the total keypoints number of ground truth, unmatched query, and unmatched database. 

The outlier rejection loss $\mathcal{L}_{or}$ serves to remove incorrect matches. 
It enhances the model's robustness against outliers and is defined as:
\begin{equation}
    \mathcal{L}_{or} = - \frac{1}{N_c}\sum_{i=1}^{N_c} w_i \left(y_i \log p_i + (1 - y_i) \log (1 - p_i) \right),
\end{equation}
where \(N_c\) means the total number of initial correspondences.
The classifier's probability output for each correspondence is indicated by $p_i$.
The target label for the correspondence is represented by $y_i$, and $w_i$ refers to the balance weight for negative and positive samples.

%% file: sec/4_experiments.tex
\section{Experiments}
\label{sec:experiments}

\begin{table*}[!t]
\begin{center}
\renewcommand\arraystretch{1.4}
\resizebox{.99\textwidth}{!}
{
\tiny
\begin{tabular}{llcccccc}
\hline \specialrule{0.5pt}{0.5pt}{0.5pt}
\multicolumn{2}{l}{\multirow{2}{*}{Methods}} &
  Reproj. AUC (\%) &
  Rotation ($^\circ$)  &
  Translation &
  \multirow{2}{*}{Time (ms) $(\downarrow)$} \\ 

\multicolumn{2}{l}{} &
  @1 / 5 / 10px ($\uparrow$) &
  \multicolumn{2}{c}{Quantile @25 / 50 / 75\% $(\downarrow)$} & 
   \\ \hline \specialrule{0.5pt}{0.5pt}{0.5pt}
\multirow{4}{*}{k=1} &
  Oracle & 34.59 / 85.02 / 92.02 &\phantom{1} 0.04 /\phantom{1} 0.06 /\phantom{1} 0.12& 0.00 / 0.01 / 0.01 &- \\
&
  GoMatch~\cite{zhou2022geometry} &
  \phantom{1} 5.67 / 22.43 / 28.01 &
  \phantom{1} 0.60 / 10.08 / 34.63 & 
  0.06 / 1.06 / 3.73 & \textbf{24.4} \\  
 &
  DGC-GNN~\cite{wang2024dgc} &
  {10.20 / 37.64 / 44.04} &\phantom{1} {0.15} /\phantom{1} {1.53} / {27.93} &
  {\textbf{0.01} / 0.15 / 3.00} & 77.8 \\ \cline{3-6}
 &
  A2-GNN & \textbf{12.72 / 41.84 / 48.02} &
  \phantom{1} \textbf{0.12} /\phantom{1} \textbf{0.79} / \textbf{26.37} &
  \textbf{0.01 / 0.08 / 2.80} & {34.0} \\\cline{1-6}
\multirow{3}{*}{k=10} &
  GoMatch~\cite{zhou2022geometry} &
  \phantom{1}8.90 / 35.67 / 44.99 &
  \phantom{1}0.18 / \phantom{1}1.29 / 16.65 &
  0.02 / 0.12 / 1.92  & \textbf{263.1} \\ 
 &
 DGC-GNN~\cite{wang2024dgc} &
  {15.30 /  51.70 /  60.01} & \phantom{1}{0.07} /\phantom{1} {0.26} /\phantom{1} {5.41} &
   \textbf{0.01 / 0.02} / 0.57 & 701.9 \\ \cline{3-6} 
 &
  \textbf{A2-GNN} &
  \textbf{17.29 / 54.41 / 62.24} &\textbf{0.06} /\phantom{1} \textbf{0.19} /\phantom{1} \textbf{4.6} &
  \textbf{0.01 / 0.02 / 0.48} & 372.0 \\ \hline \specialrule{0.5pt}{0.5pt}{0.5pt}
\end{tabular}}
\caption{Matching results on MegaDepth~\cite{li2018megadepth}. The results include the reprojection AUC, rotation and translation errors, and inference time.
Parameter $k$ is the number of retrieved images. 
The best results are bold. 
} 
\label{tab: matching}
\end{center}
\end{table*}

\subsection{Implementation Details}

\textbf{Training.} 
We train our model on the MegaDepth dataset~\cite{li2018megadepth}. The number of nearest neighbors to build the local graph is set to $k = 9$. For annular convolutions, we use $g = 3$, meaning each group contains 3 nodes. 
The model is trained using the ADAM optimizer~\cite{kingma2014adam} with a learning rate $lr = 0.001$. Training is conducted on a single 32GB Tesla V100 GPU with a batch size $b = 16$. The entire training process takes around 22 hours to complete 50 epochs.

\noindent
\textbf{Datasets.} 
Following the setting in GoMatch~\cite{zhou2022geometry} and DGC-GNN~\cite{wang2024dgc}, we train and evaluate our model on the MegaDepth dataset and conduct visual localization evaluations on Cambridge Landmark \cite{kendall2015posenet} and 7Scenes~\cite{shotton2013scene} datasets. MegaDepth is a large-scale outdoor dataset with 196 scenes captured around the world.
We train our outdoor model on 99 scenes, validate on 16 scenes and test on 53 scenes.
The ground truth sparse 3D point clouds are reconstructed by the COLMAP~\cite{schoenberger2016sfm}. 
Cambridge Landmarks is a middle-scale outdoor dataset and
7Scenes is a small indoor dataset. 
The 3D reconstruction of Cambridge Landmarks is obtained by SfM, we evaluate the localization accuracy on four out of six scenes.
The 7Scenes dataset, captured with an RGB-D camera, provides sequences where the ground truth camera poses are determined using a SLAM system, we evaluate our model on all 7Scenes as in~\cite{zhou2022geometry, wang2024dgc}.

\noindent
\textbf{Ground Truth Correspondences.} 
During training, we obtain the ground truth correspondences by reprojecting the 3D point clouds from the top-$k$ retrieved images onto the query image plane. A correspondence is considered ground truth if the reprojection error is less than 0.001 in normalized image coordinates.

\noindent
\textbf{Evaluation Metrics.} 
Similar to~\cite{zhou2022geometry,wang2024dgc}, we report the AUC score for mean reprojection error at 1 / 5 / 10 pixel and translation and rotation errors quantiles at 25 / 50 / 75\% as evaluation metrics on MegaDepth dataset.
For Cambridge and 7Scenes, we report the commonly used median translation and rotation errors per scene.
The camera pose estimation is determined by using the PnP-RANSAC~\cite{fischler1981random,gao2003complete} for the final established correspondences.
Oracle results are obtained by utilizing ground truth matches as predictions.

\begin{table*}[t!]
\begin{center}
\renewcommand\arraystretch{1.6}
\setlength{\tabcolsep}{3pt}
\resizebox{0.99\textwidth}{!}
{
\Huge

\begin{tabular}{llccccccc|ccccccccc}
\hline \specialrule{2.5pt}{0.5pt}{0.5pt}
\multicolumn{2}{l}{\multirow{2}{*}{{Methods}}} &
  \multicolumn{1}{c}{\multirow{2}{*}{{\begin{tabular}[c]{@{}c@{}}No Desc. \\ Maint.\end{tabular}}}} &
  \multicolumn{1}{c}{\multirow{2}{*}{{Privacy}}} &
  \multicolumn{4}{c}{{Cambridge-Landmarks~\cite{kendall2015posenet} (cm, $^\circ$)}} &
  \multirow{2}{*}{{MB used}} &
  {} &
  \multicolumn{7}{c}{{7Scenes~\cite{shotton2013scene} (cm, $^\circ$)}} &
  \multirow{2}{*}{{MB used}} \\ \cline{5-8} \cline{11-17}
\multicolumn{2}{l}{} &
  \multicolumn{1}{c}{} &
  \multicolumn{1}{c}{} &
  {King’s} &
  {Hospital} &
  {Shop} &
  {St. Mary’s} &
   &
  {} &
  {Chess} &
  {Fire} &
  {Heads} &
  {Office} &
  {Pumpkin} &
  {Kitchen} &
  {Stairs} &
   \\ \hline \specialrule{2.5pt}{0.5pt}{0.5pt}
 \multirow{3}{*}{\rotatebox{90}{{E2E}}} &
  {MS-Trans.~\cite{shavit2021learning}} &  \Checkmark
   &   \Checkmark
   &
  83 / 1.47 &
  181 / 2.39 &
  86 / 3.07 &
  162 / 3.99 &
  \phantom{11}71 &
   &
  11 / 4.66 &
  24 / 9.60 &
  14 / 12.19 &
  17 / 5.66 &
  18 / 4.44 &
  17 / 5.94 &
  26 / 8.45 &
  \phantom{111}71 \\
 &
  {DSAC*~\cite{brachmann2021visual}} &  \Checkmark
   &   \Checkmark
   &
  \textbf{15 / 0.30} &
  \phantom{1}21 / 0.40 &
  \textbf{\phantom{1}5 / 0.30} &
  \phantom{1}13 / 0.40 &
  \phantom{1}112 &
   &
  \phantom{1}\textbf{2} / 1.10 &
  \phantom{1}\textbf{2} / 1.24 &
  \phantom{1}\textbf{1} / 1.82 &
  \phantom{1}\textbf{3} / 1.15 &
  \phantom{1}\textbf{4} / 1.34 &
  \phantom{1}\textbf{4} / 1.68 &
  \phantom{1}\textbf{3} / 1.16 &
  \phantom{11}196 \\
 &
  {HSCNet~\cite{li2020hierarchical}} & \Checkmark
   &   \Checkmark
   &
  18 / \textbf{0.30} &
  \phantom{1}\textbf{19 / 0.30} &
  \phantom{1}6 / \textbf{0.30} &
  \phantom{11}\textbf{9 / 0.30} &
  \phantom{1}592 &
   &
  \phantom{1}\textbf{2 / 0.70} &
  \phantom{1}\textbf{2 / 0.90} &
  \phantom{1}\textbf{1 / 0.90} &
  \phantom{1}\textbf{3 / 0.80} &
  \phantom{1}\textbf{4 / 1.00} &
  \phantom{1}\textbf{4 / 1.20} &
  \phantom{1}\textbf{3 / 0.80} &
  \phantom{1}1036 \\
 \hline
 \multirow{3}{*}{\rotatebox{90}{{DB}}} & 
 {HybridSC~\cite{camposeco2019hybrid}} &  \XSolidBrush
   & --
   &
  81 / 0.59 &
  \phantom{1}75 / 1.01 &
  19 / 0.54 &
  \phantom{1}50 / 0.49 &
  \phantom{111}3 &
  \multicolumn{1}{c}{} &
  \multicolumn{1}{c}{-} &
  \multicolumn{1}{c}{-} &
  \multicolumn{1}{c}{-} &
  \multicolumn{1}{c}{-} &
  \multicolumn{1}{c}{-} &
  \multicolumn{1}{c}{-} &
  \multicolumn{1}{c}{-} &
  - \\ &
  {AS~\cite{sattler2016efficient}}&  \XSolidBrush
  &  \XSolidBrush
   &
  13 / 0.22 &
  \phantom{1}20 / 0.36 &
  \phantom{1}\textbf{4} / 0.21 &
  \phantom{11}8 / 0.25 &
  \phantom{1}813 &
   &
  \phantom{1}3 / 0.87 &
  \phantom{1}\textbf{2} / 1.01 &
  \phantom{1}\textbf{1} / 0.82 &
  \phantom{1}4 / 1.15 &
  \phantom{1}7 / 1.69 &
  \phantom{1}5 / 1.72 &
  \textbf{4} / \textbf{1.01} &
  - \\
 &
  {SP~\cite{detone2018superpoint}+SG~\cite{sarlin2020superglue}} &  \XSolidBrush
    &  \XSolidBrush
   &
  \textbf{12} / 0.20 &
  \phantom{1}15 / 0.30 &
  \phantom{1}\textbf{4 / 0.20} &
  \phantom{11}\textbf{7 / 0.21} &
  3215 &
   &
  \phantom{1}\textbf{2} / 0.85 &
  \phantom{1}\textbf{2} / 0.94 &
  \phantom{1}\textbf{1 / 0.75} &
  \phantom{1}\textbf{3} / 0.92 &
  \phantom{1}5 / 1.30 &
  \phantom{1}\textbf{4 }/ 1.40 &
  5 / 1.47 &
  22977 \\ \hline
 \multirow{3}{*}{\rotatebox{90}{{DF}}} &
  {GoMatch~\cite{zhou2022geometry}} &  \Checkmark
   & \Checkmark
   &
  25 / 0.64 &
  283 / 8.14 &
  48 / 4.77 &
  335 / 9.94 &
  \phantom{11}48 &
   &
  \phantom{1}4 / 1.65 &
  13 / 3.86 &
  \phantom{1}9 / 5.17 &
  11 / 2.48 &
  16 / 3.32 &
  13 / 2.84 &
  \ 89 / 21.12 &
  \phantom{11}302 \\
 &
  {DGC-GNN~\cite{wang2024dgc}} & \Checkmark
   &  \Checkmark
   &
  {18 / 0.47} &
  {75 / 2.83} &
  {15 / 1.57} &
  {106 / 4.03} &
  \phantom{11}69 &
   &
  \phantom{1} {\textbf{3} / 1.41} &
  \phantom{1} {\textbf{5} / 1.81} &
  \phantom{1} {\textbf{4} / 3.13} &
  \phantom{1} {7 / 1.66} &
  \phantom{1} {8 / 2.03} &
  \phantom{1} {8 / 2.14} &
  \ {83 / 21.53} &
  \phantom{11}355 \\
& 
\textbf{A2-GNN} & \Checkmark
   &  \Checkmark
   &
  \textbf{15 / 0.39} &
  \textbf{59 / 1.74} &
  \textbf{12 / 1.16} &
  \textbf{\phantom{1}76 / 2.65} &
  \phantom{11}69 &
   &
  \phantom{1}\textbf{3 / 1.37} &
  \phantom{1}\textbf{5 / 1.78} &
  \phantom{1}\textbf{4 / 2.70} &
  \phantom{1}\textbf{6 / 1.56} &
  \phantom{1}\textbf{7 / 1.86} &
  \phantom{1}\textbf{7 / 2.00} &
 \ \textbf{72 / 17.05} &
  \phantom{11}355
 \\ \hline \specialrule{2.5pt}{0.5pt}{0.5pt}
\end{tabular}
}
\caption{The comparison to existing localization baselines. E2E, DB and DF indicate end-to-end methods, descriptor-based and descriptor-free methods, respectively. 
Median translation and rotation errors for each scene are reported, as well as storage demand. The best results in each group are bold.}
\label{tab: localization}
\end{center}
\end{table*}

\subsection{Results}

\textbf{Matching Results.}
Table~\ref{tab: matching} illustrates the matching results on MegaDepth dataset using top-$1$ and top-$10$ retrieved images.
The proposed A2-GNN outperforms both GoMatch and DGC-GNN. With $k = 1$ retrieved image, A2-GNN shows significant improvements in reprojection AUC, outperforming GoMatch by 7.05 / 19.41 / 20.01\% and DGC-GNN by 2.52 / 4.2 / 3.98\%. In terms of running time, A2-GNN is slightly slower than GoMatch due to the additional encoding of RGB and geometric information, but it is x2.2 times faster than DGC-GNN by avoiding multiple times point clustering operations. When using $k = 10$ retrieved images, the same conclusion holds: A2-GNN consistently outperforms the other methods, demonstrating its robust learning capabilities.

\begin{table*}[t!]
\begin{center}

\renewcommand\arraystretch{1.4}
\setlength{\tabcolsep}{3pt}
\resizebox{0.98\textwidth}{!}
{
\begin{tabular}{lcccccccc}
\hline
\specialrule{0.5pt}{0.5pt}{0.5pt}
\multirow{2}{*}{Methods} &

\multirow{2}{*}{OR Input} &
\multicolumn{3}{c}{Self-Attention} &
\multirow{2}{*}{Color} &

  Reproj. AUC (\%) &
  Rotation ($^\circ$) &
  Translation \\
&   & Maxpooling  &  Annular & Angle & & @1 / 5 / 10px  ($\uparrow$)           & \multicolumn{2}{c}{Quantile@25 / 50 / 75\% ($\downarrow$)}        \\ \hline 
\specialrule{0.5pt}{0.5pt}{0.5pt}
GoMatch~\cite{zhou2022geometry}       & Feat. &  \Checkmark & &  &    & \phantom{1}8.90 / 35.67 / 44.99 & 0.18 / 1.29 / 16.65 & 0.02 / 0.12 / 1.92 \\ \hline 

\multirow{5}{*}{Variants}  & BV &   \Checkmark & & &   & 10.57 / 39.22 / 47.98           & 0.13 / 0.96 / 17.77           & 0.01 / 0.09 / 1.96      \\
& BV & & \Checkmark &  &    &  12.02 / 42.42 / 50.67           &  0.10 / 0.67 / 15.99         &  0.01 / 0.06 / 1.76      \\
& BV & \Checkmark & \Checkmark &  &  &  
                                        
14.04 / 47.49 / 55.82 & 0.08 / 0.40 / 11.40       & 0.01 / 0.04 / 1.20 \\        
& BV & \Checkmark & \Checkmark & \Checkmark &  &  14.79 / 49.11 / 57.41           & 0.08 / 0.33 / \phantom{1}9.14         &  0.01 / 0.03 / 1.06      \\
& Feat. &  \Checkmark & \Checkmark & \Checkmark & \Checkmark &
15.97 / 52.01 / 59.88 & 0.06 / 0.23 / \phantom{1}7.34 & 0.01 / 0.02 / 0.79
\\

\hline
\textbf{A2-GNN}     & BV & \Checkmark & \Checkmark & \Checkmark & \Checkmark & \textbf{17.29 / 54.41 / 62.24} & \textbf{0.06 / 0.19 / \phantom{1}4.60} & \textbf{0.01 / 0.02 / 0.48} 

\\ \hline
\specialrule{0.5pt}{0.5pt}{0.5pt}
\end{tabular}
}
\caption{Architecture Ablations. The reprojection AUC, rotation and translation errors are reported. OR Input indicates the input types of outlier rejection. Feat. means enhanced feature from GNN as input, and BV refers to bearing vector. The best results are bold.
} 
\label{tab:abla}
\end{center}
\end{table*}

\noindent
\textbf{Visual Localization.} 
As presented in Table~\ref{tab: localization}, A2-GNN outperforms GoMatch and DGC-GNN, achieving state-of-the-art results in the visual descriptor-free group. Specifically, the average pose error for the Cambridge Landmarks dataset is 40.5 cm / 1.48$^\circ$ for A2-GNN, compared to 54 cm / 2.23$^\circ$ for DGC-GNN and 173 cm / 5.87$^\circ$ for GoMatch. For the 7Scenes dataset, the average pose error is 14 cm / 4.04$^\circ$ for A2-GNN, 16 cm / 4.81$^\circ$ for DGC-GNN, and 22 cm / 5.77$^\circ$ for GoMatch. Additionally, A2-GNN retains the advantages of DF methods, including low storage requirements, privacy preservation, and low maintenance costs.

\noindent
\textbf{Generalizability.}
Similar to previous work~\cite{zhou2022geometry,wang2024dgc}, we evaluate the generalizability of our model across different datasets and keypoint detectors for visual localization tasks. Our model is trained on the MegaDepth~\cite{li2018megadepth} using the SIFT~\cite{lowe2004distinctive} detector and evaluated on the indoor 7Scenes dataset~\cite{shotton2013scene} with the SIFT~\cite{lowe2004distinctive} keypoint detector and the outdoor Cambridge dataset~\cite{kendall2015posenet} using the SuperPoint~\cite{detone2018superpoint} detector. These experiments are summarized in Table~\ref{tab: localization}, providing a comprehensive overview of our A2-GNN performance under varying training and evaluation conditions. Remarkably, even when using the SuperPoint detector on the Cambridge dataset, A2-GNN demonstrates strong performance, highlighting its robust generalizability. 
\begin{figure}[!t]
  \centering
  \includegraphics[width=\linewidth]{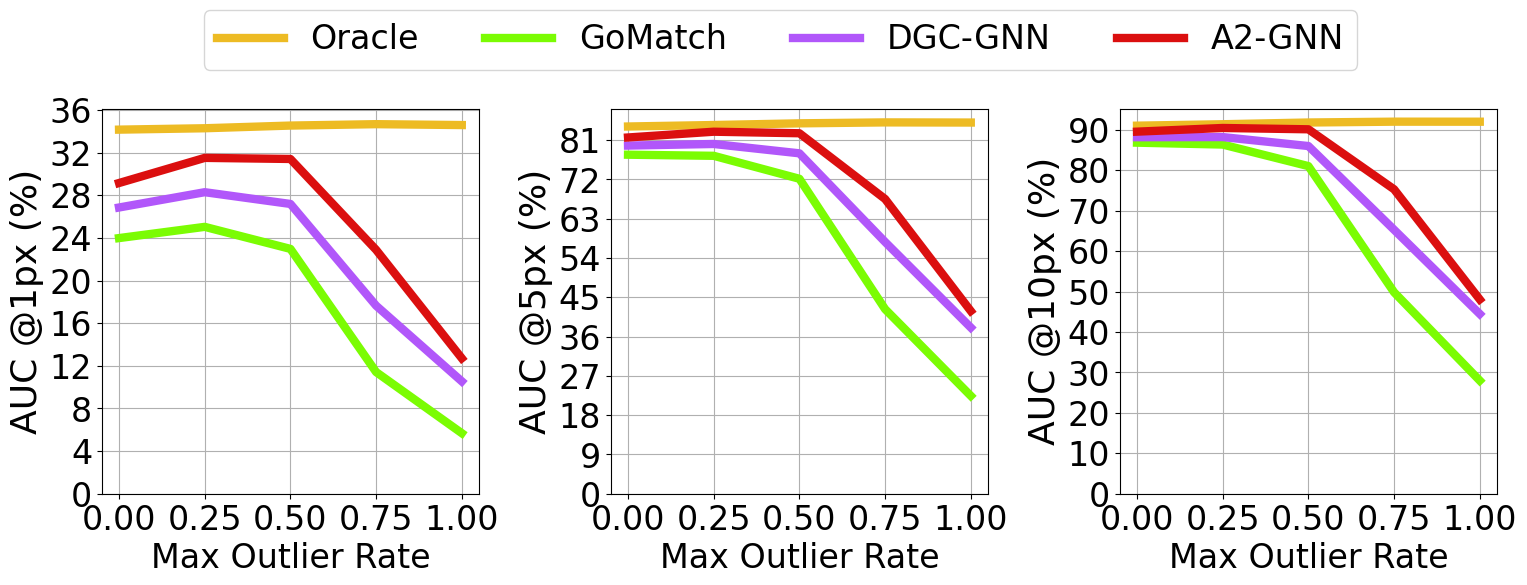}
  \caption{Outlier Sensitivity.  Comparison of the AUC for GoMatch~\cite{zhou2022geometry}, DGC-GNN~\cite{wang2024dgc}, and the proposed A2-GNN under different outlier ratios at 1, 5, and 10 pixels thresholds.
  Oracle is the upper bound by using ground truth matches.
  }
  \label{fig: outliers}
\end{figure}

\noindent
\textbf{Ablation Studies.}
We validate our A2-GNN module design by conducting ablation studies on the MegaDepth dataset~\cite{li2018megadepth} with the top-10 retrieved images ($k=10$). The ablation results are presented in Table~\ref{tab:abla}. First, we analyze the impact of the outlier rejection inputs by changing the input from enhanced features to bearing vectors. This change improves the reprojection AUC by 1.67 / 3.55 / 2.99 \%, confirming our assumption that explicit positional information can enhance consensus in epipolar geometry, thereby improving outlier rejection accuracy. 
Next, we perform ablations on our geometric feature embedding. Incorporating both max-pooling and annular convolution significantly boosts reprojection AUC by 3.47 / 8.27 / 7.84 \%. 
We can conclude that max-pooling operation holds the permutation invariance, a property that is vital for robust feature learning.
Additionally, angle embedding further enhances geometric embedding and shows improvement by 0.75 / 1.62 / 1.59\%. 
Similar to~\cite{wang2024dgc}, we also observe notable gains when color information is added to the point feature.

The evaluation results without outlier rejection are shown in Table~\ref{tab: cl}. We use image retrieval with $k=1$ and set the outlier rejection threshold to 0. Our A2-GNN demonstrates the ability to generate high-quality 2D-3D correspondences without outlier rejection, outperforming both GoMatch and DGC-GNN by large margins.  
Unlike GoMatch and DGC-GNN, A2-GNN uses geometric features refined by GNN layers solely for matching, rather than simultaneously for outlier rejection. This highlights the goal disparities between matching and removing outliers.

\begin{table}[t!]

\centering
\renewcommand\arraystretch{1.5}
\Large
\resizebox{.48\textwidth}{!}{
  \begin{tabular}{lccc}
    \toprule
    \specialrule{0.5pt}{0.5pt}{0.5pt}
    \multirow{2}{*}{Method}   & Reproj. AUC (\%) &
  Rotation ($^\circ$)  &
  Translation (m)  \\ 
  & @1 / 5 / 10px ($\uparrow$) &
  \multicolumn{2}{c} {Quantile @25 / 50 / 75\% $(\downarrow)$} \\
    \midrule
    GoMatch (no OR) & \ 4.47 / 17.95 / 23.42 &1.29 / 11.85 / 33.60&0.11 / 1.18 / 3.58 \\
    DGC-GNN (no OR) & \ 8.56 / 30.79 / 37.02& 0.22 /  \ 4.85 / 30.07 & 0.02 / 0.47 / 3.10 \\
    \textbf{A2-GNN (no OR)} & \textbf{11.37 / 37.04 / 43.15 }&  \textbf{ 0.13 / \ 2.32 / 27.00} & \textbf{0.01 / 0.22 / 2.87 }\\
    \bottomrule
    \specialrule{0.5pt}{0.5pt}{0.5pt}
\end{tabular}
}
\caption{Ablation results without outlier rejection on MegaDepth on top-1 image retrieval. 
} 
\label{tab: cl}
\end{table}

\noindent
\textbf{Sensitivity to Keypoint Outliers.}
To evaluate our method's performance in handling keypoint outliers, we follow the procedure outlined in~\cite{zhou2022geometry, wang2024dgc} and vary the outlier ratios from 0 to 1 during inference. We present the AUC across various pixel thresholds and outlier ratios in Fig.~\ref{fig: outliers}. When the outlier ratio is 0, all input keypoints are derived from the ground truth, meaning no outliers are present. Conversely, at a ratio of 1, all keypoints are directly taken from the 2D query and the 3D points from the top-k retrieved images.

As shown in Fig.~\ref{fig: outliers}, our A2-GNN model consistently outperforms GoMatch and DGC-GNN across all outlier rate scenarios. When the outlier rate is below 50\%, the AUC of our method closely approaches the Oracle upper bound.
However, when the outlier rate exceeds 50\%, performance declines rapidly. This is due to the increased difficulty in identifying correct matches in high-outlier scenarios. Additionally, our model was trained under conditions with a 50\% outlier rate and thus does not perform as well when inputs are chosen randomly.

%% file: sec/5_conclusion.tex
\section{Conclusion}

This paper introduces A2-GNN, a novel local graph neighbor aggregation method for direct 2D-3D matching without relying on visual descriptors. 
Our angle-annular convolution method effectively captures more robust local geometric information. 
Compared to the state-of-the-art descriptor-free matcher DGC-GNN~\cite{wang2024dgc}, A2-GNN demonstrates superior accuracy and efficiency, leading to significantly improved localization performance. 
These advancements help bridge the gap between descriptor-based and descriptor-free methods while addressing the limitations of descriptor-based approaches, such as high storage requirements, maintenance complexity, and privacy concerns. Overall, A2-GNN represents a substantial advancement in the field of direct 2D-3D matching.

\noindent
\textbf{Limitations.} 
While the proposed A2-GNN method demonstrates significant efficiency and accuracy improvements over existing descriptor-free approaches, its performance still lags behind that of traditional descriptor-based algorithms. This shortfall is primarily due to noisy keypoint inputs from query images and the limited information available. Specifically, when image keypoints contain more than 50\% outliers, the performance drops dramatically. Although A2-GNN advances the state of descriptor-free methods, further research is needed to bridge this gap.

\noindent
\textbf{Acknowledgements.} 
We thank the Research Council of Finland for funding the projects EnergyEff (353138), BERMUDA (362407) and PROFI7 (352788), as well as the Finnish Doctoral Program Network in Artificial Intelligence(AI-DOC, decision number VN/3137/2024-OKM-6).
We acknowledge the computational resources provided by the CSC-IT Center for Science, Finland.

%% file: sec/X_suppl.tex
\maketitlesupplementary

\setcounter{page}{1}

\section{Additional Details and Results}
\label{sec: supplementaty exp}

\textbf{Data Preparation.}
Following the experimental settings in GoMatch~\cite{zhou2022geometry} and DGC-GNN~\cite{wang2024dgc}, we use the MegaDepth dataset~\cite{li2018megadepth} for training. MegaDepth is a large-scale outdoor dataset comprising 196 scenes from various landscapes around the world. We utilize 99 scenes for training, 16 scenes for validation, and 53 scenes for testing. The ground truth sparse 3D point clouds are reconstructed using COLMAP~\cite{schoenberger2016sfm}. During data preprocessing, a maximum of 500 query images are selected per scene. For each query, we gather its $k$ co-visible views, ensuring at least 35\% visual overlap. Queries lacking sufficient co-visible views are excluded from the training set. Visual overlap is computed as the ratio of co-observed 3D points to the total number of 3D points in the query image. The training set comprises 25,624 queries from 99 scenes, the validation set includes 3,146 queries from 16 scenes, and the test set consists of 12,399 samples from 49 scenes. For the training dataset, we control the number of keypoints per image to range from 100 to 1,024. During inference, this range is adjusted to 10 to 1,024 keypoints per image.

\noindent
\textbf{Representation Ablation Study.}
We present results with different 3D representations, as shown in Table~\ref{tab: representation learning}. The bearing vector as the representation in 3D side plays a crucial role in enhancing the results. The insight behind this improvement is that it integrates the pose of database images into feature learning, bringing one step further towards middle representation from 3D to 2D.

\noindent
\textbf{Generalizability.} 
Our model is trained on the MegaDepth dataset~\cite{li2018megadepth} using the SIFT~\cite{lowe2004distinctive} detector. To demonstrate the generalizability of our model, we conducted evaluations on the 7Scenes dataset using two keypoint detectors: SIFT and SuperPoint~\cite{detone2018superpoint}. The results are presented in Table~\ref{tab:detector}. The similar results in translation and rotation errors between the two detectors further demonstrate the robustness and generalizability of our model.

\noindent
\textbf{Hyperparameters Selection.}
Ablation studies on various hyperparameters in the self-attention layer are presented in Table~\ref{tab:hyparam_abla}. The outlier rejection threshold of $t=0.7$ yields the best results, achieving higher AUC and lower rotation and translation errors. We select $t=0.5$ in the main paper to make a fair comparison with other methods. The choice of the nearest neighbors parameter $k$ has minimal impact on performance. However, when fewer nearest neighbors are processed, it becomes more challenging to accurately capture the local geometric structures.

\begin{table}[t!]

\centering
\renewcommand\arraystretch{1.5}
\Large
\resizebox{0.45\textwidth}{!}{
  \begin{tabular}{cccc}
    \toprule
    \specialrule{0.5pt}{0.5pt}{0.5pt}
    \multirow{2}{*}{3D representation}   & Reproj. AUC (\%) &
  Rotation ($^\circ$)  &
  Translation (m)  \\ 
  & @1 / 5 / 10px ($\uparrow$) &
  \multicolumn{2}{c} {Quantile @25 / 50 / 75\% $(\downarrow)$} \\
    \midrule
    Coordinate & 7.69 / 27.96 / 32.82 & 0.28 / 12.6 /  59.64 & 0.02 / 1.32 / 5.34 \\
    Bearing vector & \textbf{12.72 / 41.84 / 48.02} &
  \textbf{0.12} / \phantom{1}\textbf{0.79} / \textbf{26.37} &
  \textbf{0.01 / 0.08 / 2.80} \\
    \bottomrule
    \specialrule{0.5pt}{0.5pt}{0.5pt}
\end{tabular}
}
\caption{Ablation results with different 3D representations on MegaDepth on top-1 image retrieval. 
} 
\label{tab: representation learning}
\end{table}

\begin{table}[!t]
    \centering
    \begin{tabular}{lcc}
        \specialrule{0.5pt}{0.5pt}{0.5pt}
        \hline
         
         7Scenes~\cite{shotton2013scene} & SIFT~\cite{lowe2004distinctive} & SuperPoint~\cite{detone2018superpoint} \\
         \specialrule{0.5pt}{0.5pt}{0.5pt}
         \hline
         Chess      & 3 /\phantom{1}1.37 & 3 /\phantom{1}1.41 \\
         Fire       & 5 /\phantom{1}1.78 & 6 /\phantom{1}1.99\\
         Heads      & 4 /\phantom{1}2.70 & 2 /\phantom{1}3.12\\
         Office     & 6 /\phantom{1}1.56 & 6 /\phantom{1}1.48\\
         Pumpkin    & 7 /\phantom{1}1.86 & 9 /\phantom{1}2.28\\
         Redkitchen & 7 /\phantom{1}2.00 & 8 /\phantom{1}2.08\\
         Stairs     & 72 /17.05 & 66 / 16.02\\
         \specialrule{0.5pt}{0.5pt}{0.5pt}
         \hline

    \end{tabular}
    \caption{Comparison on sift and superpoint as detector on 7Scenes dataset. Median translation and  rotation errors ($cm,^\circ$) are reported.}
    \label{tab:detector}
\end{table}


\noindent
\textbf{Timing and Model Size.}
The inference time per query image for A2-GNN is $\sim$34 ms, comprising four main components: the feature encoding ($\sim$2 ms), the attention layers ($\sim$14 ms), optimal transport ($\sim$9 ms), and the outlier rejection process ($\sim$6 ms). Our model contains 2.7 million parameters, with a total size of $\sim$10.6 MB. All experiments were conducted on a 32GB NVIDIA Tesla V100 GPU, using a maximum of 1,024 keypoints.


\begin{table*}[h]
\begin{center}
\renewcommand\arraystretch{1.3}
\setlength{\tabcolsep}{3pt}
\resizebox{.99\textwidth}{!}
{
\begin{tabular}{lcccccc}
\hline
\specialrule{1.5pt}{0.5pt}{0.5pt}
\multirow{2}{*}{Methods}     & \multirow{2}{*}{Neighbors} & \multirow{2}{*}{Groups} & \multirow{2}{*}{OR Threshold} & Reproj. AUC (\%)  & Rotation ($^\circ$)  & Translation                 \\ & &  & & @1 / 5 / 10px  ($\uparrow$)    & \multicolumn{2}{c}{Quantile@25 / 50 / 75\% ($\downarrow$)} \\ \specialrule{1.5pt}{0.5pt}{0.5pt}
A2-GNN                      & 9      & 3                         & 0.5                           & 17.29 / 54.41 / 62.24          & \textbf{0.06} / 0.19 / 4.6 \phantom{1} & \textbf{0.01} / {0.02} / 0.48    \\ \hline
\multirow{5}{*}{HyperParam.} & 9 & 3 & 0.7                           & \textbf{18.59 / 58.84 / 66.48}        & \textbf{0.06 / 0.16 / 2.16}  & \textbf{0.01 / 0.01 / 0.22} \\    &
9 & 3  & 0.3  & 14.82 / 48.57 / 56.7 \phantom{1} &0.07 / 0.34 / 8.42& \textbf{0.01} / 0.03 / 0.9  \phantom{1}        \\ \cline{2-7} 
& 12  & 3       & 0.5        &   17.21 / 54.36 / 62.18            &  \textbf{0.06} / 0.2 \phantom{1} / 4.45  &  \textbf{0.01} / 0.02  / 0.46\\
  & 9    & no groups & 0.5     &  15.35 / 49.81 / 57.64  &   0.08 / 0.28 / 5.48    &   \textbf{0.01} / 0.03 / 0.58 \\
  & 6 & 3 & 0.5 & 16.41/ 51.84 / 59.52 & 0.06 / 0.22 / 7.49 & \textbf{0.01} / 0.02 / 0.81
        \\ \specialrule{1.5pt}{0.5pt}{0.5pt}
\end{tabular}
}
\vspace{10pt}
\caption{Ablations on Hyperparameters. 
The results are evaluated on MegaDepth with top-$10$ retrieval images. 
The best results are bold.
}
\label{tab:hyparam_abla}
\end{center}
\end{table*}
